\pdfoutput=1

\documentclass[11pt]{article}

\usepackage[review]{ACL2023}

\usepackage{times}
\usepackage{latexsym}

\usepackage[T1]{fontenc}

\usepackage[utf8]{inputenc}

\usepackage{microtype}

\usepackage{inconsolata}

\usepackage{graphicx}

\title{dIR--Discrete Information Retrieval: \\
Conversational Search over Unstructured (and Structured) Data \\
with Large Language Models}

\author{Pablo Rodriguez Bertorello \and Jean Rodmond Junior Laguerre \\
        Computer Science Department \\ Stanford University}

\begin{document}

\nolinenumbers

{\makeatletter\acl@finalcopytrue
  \maketitle
}

\begin{abstract}
Data is stored in both structured and unstructured form. Querying both, to power natural language conversations, is a challenge.

This paper introduces dIR, Discrete Information Retrieval, providing a unified interface to query both free text and structured knowledge. Specifically, a Large Language Model (LLM) transforms text into expressive representation. After the text is extracted into columnar form, it can then be queried via a text-to-SQL Semantic Parser, with an LLM converting natural language into SQL. Where desired, such conversation may be effected by a multi-step reasoning conversational agent.

We validate our approach via a proprietary question/answer data set, concluding that dIR makes a whole new class of queries on free text possible when compared to traditionally fine-tuned dense-embedding-model- based Information Retrieval (IR) and SQL-based Knowledge Bases (KB).  For sufficiently complex queries, dIR can succeed where no other method stands a chance.
\end{abstract}

\section{Introduction}\label{sec:introduction}
Many data are stored in both structured and unstructured form: product databases, patient records, and customer reviews, to name a few. This paper sets forth a novel approach for conversational agents to interact with such data, particularly taking advantage of free text information.

\begin{figure}
\caption{Products table row with both structured and unstructured data}
\label{productexample}
\centering
\includegraphics[scale=0.8, angle=0]{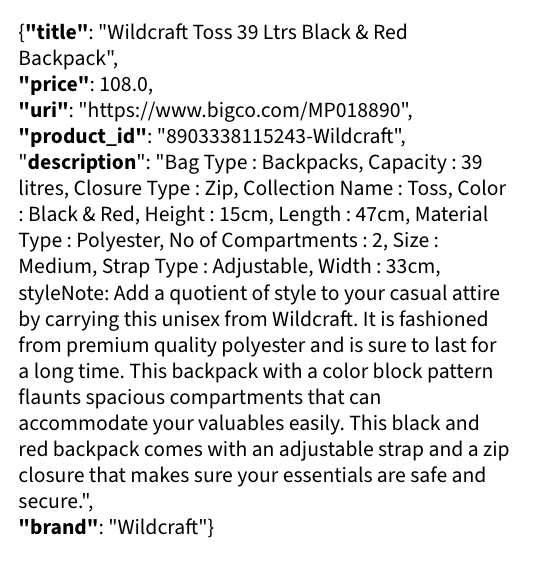} 
\end{figure}

Figure \ref{productexample} shows a data row to motivate this paper. It describes a product, with information such as title, price, and description.  To answer a question like "Do you have a non-black 15-liter backpack under \$400?", an agent needs to take into account both the structured column price, and the unstructured free text column description.

\subsection{Current Approaches}\label{sec:approaches}
Prior work has focused on question-answering on structured databases, via SQL, which is limited to basic pattern matching on strings. Thus, largely applicable to searches not needing to rely on free text information (\citealp{paper-guo-gao}; \citealp{wang-etal-2020-rat}; \citealp{scholak-etal-2021-duorat}; \citealp{DBLP:journals/corr/abs-1709-00103}). Recent approaches are based on LLMs (\citealp{hu2022incontext}; \citealp{an2023skillbased}; \citealp{nan2023enhancing}; \citealp{DBLP:journals/corr/abs-2201-11227}; \citealp{arora2023adapt})

A separate research direction focuses on information retrieval over free text corpus (\citealp{yang-etal-2019-end-end}; \citealp{DBLP:journals/corr/abs-2009-00914}; \citealp{nie-etal-2019-revealing}). More recently investigation seeks to leverage LLMs (\citealp{jiang2023active}; \citealp{semnani-etal-2023-wikichat}; \citealp{gao2023enabling}; \citealp{khattab2023demonstratesearchpredict}).

\subsection{Contributions} \label{paper-contributions}
First, \textbf{We make it possible for queries to take full advantage of hybrid sources}. That is, generate responses from free text in addition to structured conversational knowledge. At data preparation time, free text sources are used to generate structured columnar representations, via few-shot LLM prompting, as shown in Figure \ref{few-shot-colum-generation}. At query time, natural language questions are converted via few-shot LLM text-to-SQL Semantic Parsing, as shown in Figure \ref{few-shot-semantic-prompting}. Thus, no model fine-tuning is required, and neither is any modification of the SQL language standard.

Second, \textbf{We validate our dIR approach via a cross-domain proprietary question/answer data set}. This includes exploratory queries, as in Figure \ref{query-exploratory}, and complex direct queries, as in Figure \ref{query-direct}, as well as multi-hop queries. Given the arbitrary complexity of the queries studied, SQL stores string functions and vector databases can't but produce both low Query Recall and Query Precision.

\section{Related Work}\label{related-work}

\subsection{Conversational Agents}\label{conversational-agents}
Recent progress in conversational agents leverages LLMs to reason and act over data (\citealp{paper-react}; \citealp{lei2023s3hqa}; \citealp{kumar-etal-2023-multi}; \citealp{lee-etal-2023-mafid}). Such reasoner modules are capable of generating a final answer after multiple data consideration steps.  

A prior line of work utilized hybrid data selectors. However, they focused on single-turn question-answering, see Section \ref{hybrid-qna}.

\subsection{Information Retrieval}\label{information-retrieval}
The Information Retrieval (IR) field has seen the emergence of many neural ranking models that operate on dense embeddings (\citealp{paper-colbert}; \citealp{DBLP:journals/corr/abs-1711-08611}; \citealp{Xiong2017ConvolutionalNN}; \citealp{DBLP:journals/corr/abs-1903-07666}; \citealp{DBLP:journals/corr/abs-1903-06902}). Previously, ranking models relied on hand-crafted features. Still today, search matching employs embedding-based representations of queries and documents.

Recent progress in Natural Language Understanding highlights the importance of pre-training language representation models in an unsupervised fashion, before subsequently fine-tuning them on downstream tasks such as ranking. (\citealp{DBLP:journals/corr/abs-1810-04805}; \citealp{DBLP:journals/corr/abs-1904-07094}; \citealp{DBLP:journals/corr/abs-1901-04085}).

\subsection{Hybrid Question Answering} \label{hybrid-qna}
Hybrid data querying often follows the retriever-reasoner paradigm, where retriever modules are trained to identify relevant subsets of structured or unstructured data. 

\citealp{paper-suql} proposes a formal augmentation of the SQL language standard with free-text primitives. This augmentation allows the query language compiler to generate temporary tables, where columnar fields are extracted from text. Then LLM-based Text-to-SQL semantic parsing is utilized to answer queries.

\citealp{chen-etal-2020-hybridqa} focuses on single-turn question-answering, with the Hybrid QA data set for evaluation, with only 16 rows per table on average.

\subsection{Text-to-SQL Semantic Parsing}\label{semantic-parsing}
\medskip
Semantic parsing aims to translate natural language utterances into executable programs, an active research (\citealp{DBLP:journals/corr/abs-1207-1420}; \citealp{kwiatkowski-etal-2011-lexical}; \citealp{pasupat-liang-2015-compositional}; \citealp{wang-etal-2015-building}; \citealp{xu-etal-2020-autoqa}; \citealp{yu-etal-2019-sparc}). 

Relational database SQL has been widely adopted as the target executable by previous works due to its popularity. Text-to-SQL may be used for single-turn question-answering (\citealp{paper-text2sql}; \citealp{guo2023prompting}; \citealp{DBLP:journals/corr/abs-1911-04942}; \citealp{scholak-etal-2021-duorat}; \citealp{DBLP:journals/corr/abs-1709-00103}) or multi-turn conversations (\citealp{DBLP:journals/corr/abs-1909-05378}; \citealp{DBLP:journals/corr/abs-1911-04942}; \citealp{liu-etal-2022-augmenting-multi}). 

Recently, LLMs have become capable text-to-SQL semantic parsers via in-context learning (\citealp{DBLP:journals/corr/abs-2005-14165}; \citealp{guo2023prompting}), following a variety of prompts (\citealp{hu2022incontext}; \citealp{DBLP:journals/corr/abs-2201-11227}; \citealp{an2023skillbased}; \citealp{nan2023enhancing}; \citealp{arora2023adapt}; \citealp{guo2023prompting}; \citealp{sun2023sqlprompt}; \citealp{zhang2023reactable}).

\section{Core Ideas} \label{core-ideas}
Underlying traditional Information Retrieval per Section \ref{information-retrieval}, often the Approximate Nearest Neighbors algorithm is utilized. They arrive at a dense embedding representation text in aggregate, for the entire query or document chunk. Thus, it fails to take advantage of implicit regularities in the text, where they exist.

These regularities often exist in unstructured text. Consider product databases as exemplified in Figure \ref{productexample}, patient records, and restaurants, to name a few. For a backpack product table, the description text field likely indicates the backpack size or handle type, among several other facts. For a patient record, a description field likely utilizes anatomy terminology. For a restaurant, there is some likelihood that the customer review field likely indicates whether it is pet-friendly.

Hybrid question-answering approaches per Section \ref{hybrid-qna}, often powered by Information Retrieval, inherit IR's weaknesses. Not only do they require domain fine-tuning embedding models. They also result in low complex query recall and low complex query precision. While IR may be able to tell apart a backpack in a table of watches, it is most likely unable to rank the correct backpack rows for a search requiring that the backpack's size and strap type be considered, as that may be a small part of the text description field. 

This paper sets forth a novel approach for conversational agents to interact with unstructured in addition to structured data: 

\subsection{Taking advantage of free text information regularity} \label{discretize-step}

\begin{itemize}
   \item \textbf{Context:} consider a table like the one in Figure \ref{productexample}, containing a primary key, in this case product\_id
   \item \textbf{Inference:} let a new table, with the same primary key
   \item \textbf{Collect:} identify the text fields from the Context that should be utilized in the Discretize step 
   \item \textbf{Discretize:} for the text from the Collect step, run LLM inference, per the prompt in Figure \ref{few-shot-colum-generation}. The result is a list of key-value tuples, where the key is a categorization, and the value is the enumerated value for that category.  For example ('product\_size', '15 liter'), ('handle\_type', 'strap')
   \item \textbf{Enumerate:} reduce the categorization from the Discretize step into enumerated types like \{ 'product\_size': ['15 liter', '22 liter'] \}
   \item \textbf{Generate table:} for each key from the Enumerate step, generate a column in the Inference table. 
\end{itemize}

Several of these steps are objectively optimizable, which is outlined as future work in Section \ref{future-work}.

\subsection{Providing a unified SQL interface} \label{unified-interface}

Once the process in Section \ref{discretize-step} is completed:
\begin{itemize}
   \item \textbf{(Optional) Execution:} let a ReAct \cite{paper-react} reasoner, which in response to a user question executes a cycle of:
      \subitem \textbf{Thought:} chain-of-thought prompting  
      \subitem \textbf{Action:} action plan generation, for instance converting the user's intention into a tool call and query arguments, to fulfill the following
      \subitem \textbf{Observation:} returning the tool's hybrid structured/unstructured query results for further thought/action to conclusion by the agent 
    \item \textbf{Text-to-SQL:} a user or tool's query is converted to SQL, which is run against the JOIN of the Context table and Inference tables created above
\end{itemize}

As the user query is converted into SQL queries over a table that includes pre-existing structure and fields extracted from text, it returns maximum insight from the data.

\section{Experimental Results} \label{experimental-results}

The evaluation was conducted on a proprietary data set consisting of 33 different sub-domains, totaling 11,967 products (over 362 products on average for each). That is over 20 times more rows per table than in the HybridQA dataset.

The main types of evaluations were:
\begin{itemize}
   \item \textbf{Direct Query:} where the natural language question expresses very specific user interest, as in Figure \ref{query-direct}
   \item \textbf{Exploratory Query:} where the natural language question expresses a vague user interest, as in Figure \ref{query-exploratory}. The level of complexity in queries simply escapes what is possible with unstructured Information Retrieval or structured SQL on text fields
   \item \textbf{Multi-hop Query:} where the natural language question requires reasoning over multiple hybrid query responses, which will be published in a separate paper
\end{itemize}

We estimated the relative performance of different LLMs:
\begin{itemize}
   \item \textbf{GPT 3.5:} which was sufficiently effective for the Discretize step
   \item \textbf{GPT 4.0:} which we found most effective for the text-to-SQL step, while being unnecessarily expensive for the Discretize step
   \item \textbf{Palm2:} which was neither the best at Discretize nor at text-to-SQL
\end{itemize}

Per our analysis, one of the critical trade-offs was whether to put all 33 sub-domains of product data into a single Inference database table. Since we sought to answer the most complex natural language questions possible that needed to be resolved via fields extracted from free text, we opted to put each sub-domain in its separate table. 

\section{Limitations}
The Discretize step could generate thousands of columns, even for a data sub-domain, which is to be optimized:
\begin{itemize}
   \item \textbf{SQL system limitations:} databases support a maximum number of columns per table. At present that ranges from 2048 columns for SQL Lite, to 4096 columns for MySQL, to 20480 for Spark
   \item \textbf{LLM inference limitations:} the greater the number of column-value pairs identified in Enumerate, the larger the input context that needs to be passed at text-to-SQL inference time. At present, input character limits range from 4096 tokens for GPT 3.5, to 32k tokens for GPT 4.0, to 128k tokens for GPT 4 Turbo. Even if enumerations could fit in the context for a cross-domain query, LLM inference cost is on a per token basis and thus should be minimized
\end{itemize}

In our experiments, we capped the number of fields extracted by keeping column name complexity to a minimum.  For example, product\_brand was kept, while number\_of\_pockets was discarded. This was sufficient for enumeration to fit within the column limit of SQL Lite and the token limit of GPT 4.0.

Importantly, we noted the importance of cross-domain grounding in generated SQL, which we accomplished by ensuring that some enumerated fields were extracted in every sub-domain, for example: product\_type

At scale, several of these aspects may be rigorously optimized, which we leave for future work per Section \ref{future-work}.

\section{Future Work} \label{future-work}
First, housing each sub-domain of data in separate tables creates the complexity that a conversational agent then needs to be able to take a query Action that is specific to a particular table. The alternative would be too expensive, to run separate LLM inferences in each domain. A possible path forward would be to rely on a separate model to direct the conversational agent's query to the best data source table. This may be where ReAct should be combined with Reinforcement Learning to further unlock the conversational potential of Large Language Models.

Second, the following algorithmic steps may be objectively optimizable:
\begin{itemize}
   \item \textbf{Grounding:} when natural language text is converted to SQL, the few-shot prompt provided to the LLM includes Enumerate information, per Figure \ref{few-shot-semantic-prompting}. However, a question about backpacks directed to a table about perfumes could result in undesired results, unless cross-table/domain grounding is provided
   \item \textbf{Discretize:} it may be possible to optimize the few-shot prompt provided to the LLM, such that the extracted database fields provide the best coverage of all data sub-domains, as opposed to just being good for each one row individually 
   \item \textbf{Enumerate:} leveraging LLMs, it may be possible to best linguistically reduce different enumerated keys.  For example, these two fields could be consolidated into one: no\_of\_pockets, and number\_of\_pockets
   \item \textbf{Domains:} whether different data domains could or should be stored in shared tables, as a function of the depth of querying desired, while constrained by the maximum number of columns per table that is supported by the database system of choice
\end{itemize}

Finally, we demonstrated that query semantic parsing may be utilized to establish a Dialog State. Further research should be conducted on a conversational agent utilizing this state information to best understand a user's intentions as it evolves over multi-step dialog. See Figure \ref{query-exploratory} and Figure \ref{query-direct}.

\section{Conclusion} \label{the-conclusion}
We introduce dIR, Discrete Information Retrieval, a method and process providing a unified interface to query both free text and structured knowledge. It leverages LLMs to transform free text into an expressive structured representation, which can be queried in natural language, via a text-to-SQL Semantic Parser, with an LLM converting natural language into SQL. Where desired, such conversation may be effected by a multi-step reasoning conversational agent. Our validation with a proprietary cross-domain dataset reveals that dIR makes a whole new class of arbitrarily complex queries possible on information that may originally be stored as free text. While several aspects should be optimized in the future, it is performant at this onset, without requiring LLM fine-tuning nor a modification of the SQL language standard.

\bibliography{anthology,custom}
\bibliographystyle{acl_natbib}

\appendix 

\section{Appendix}

\subsection{Query Complexity} \label{sec:appendix}
\begin{figure}
\caption{Exploratory Query: Response and Dialog State}
\label{query-exploratory}
\centering
\includegraphics[scale=0.8, angle=0]{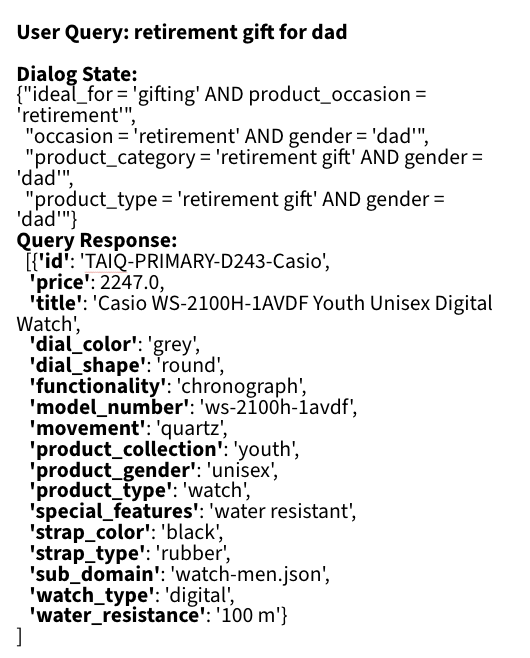} 
\end{figure}

\begin{figure}
\caption{Direct Query: Response and Dialog State}
\label{query-direct}
\centering
\includegraphics[scale=0.8, angle=0]{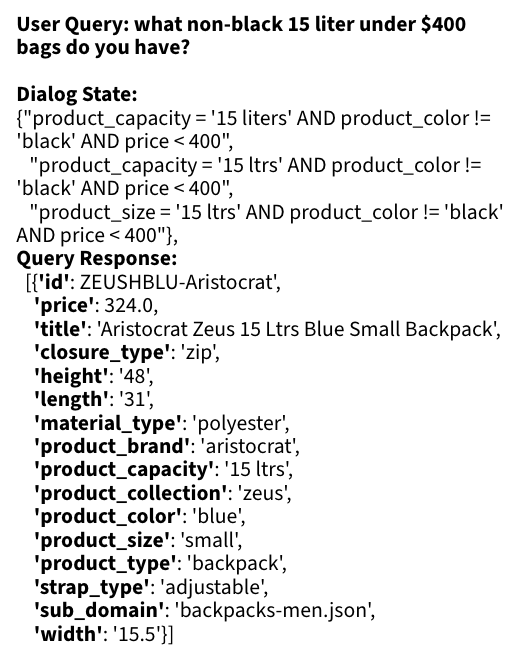} 
\end{figure}

The key queries studied are Exploratory as in Figure \ref{query-exploratory}, and Direct as in Figure \ref{query-direct}.

\subsection{LLM Prompt} \label{llm-prompts}
\begin{figure}
\caption{LLM Prompt to Generate Structured Columnar Representations from Free Text}
\label{few-shot-colum-generation}
\centering
\includegraphics[scale=0.60, angle=0]{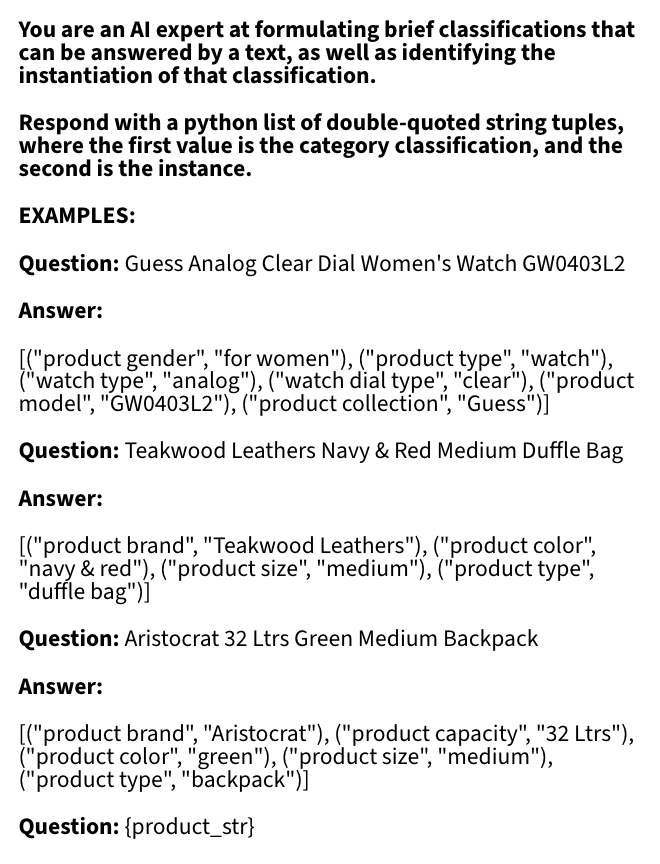} 
\end{figure}

\begin{figure}
\caption{LLM Prompt for Text-to-SQL Semantic Parsing}
\label{few-shot-semantic-prompting}
\centering
\includegraphics[scale=0.60, angle=0]{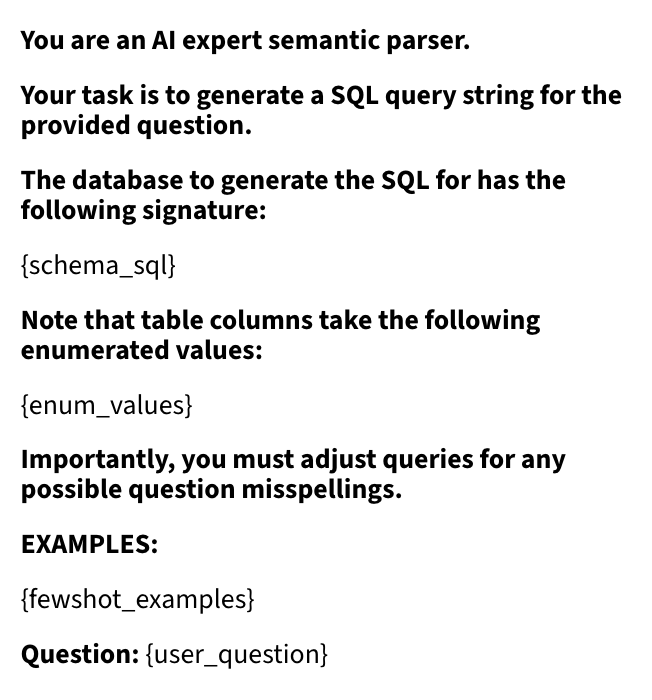} 
\end{figure}

The fundamental LLM prompts developed are Generate as in Figure \ref{few-shot-colum-generation}, and Text-toSQL as in Figure \ref{few-shot-semantic-prompting}.

\end{document}